\documentclass[10pt,twocolumn,letterpaper]{article}
\usepackage{times}
\usepackage{epsfig}
\usepackage{graphicx}
\usepackage{amsmath}
\usepackage{amssymb}
\usepackage{multirow}
\usepackage{tabularx}
\usepackage{array}
\usepackage{tabulary}
\usepackage{makecell}

\usepackage[breaklinks=true,bookmarks=false]{hyperref}

\begin{document}

\title{Exploiting Web Images for Weakly Supervised Object Detection}

\author{Qingyi Tao\qquad
Hao Yang
\qquad Jianfei Cai\\
Nanyang Technological University, Singapore
}
\date{}


\maketitle

\begin{abstract}
In recent years, the performance of object detection has advanced significantly with the evolving deep convolutional neural networks. However, the state-of-the-art object detection methods still rely on accurate bounding box annotations that require extensive human labelling. Object detection without bounding box annotations, i.e, weakly supervised detection methods, are still lagging far behind. As weakly supervised detection only uses image level labels and does not require the ground truth of bounding box location and label of each object in an image, it is generally very difficult to distill knowledge of the actual appearances of objects. Inspired by curriculum learning, this paper proposes an easy-to-hard knowledge transfer scheme that incorporates easy web images to provide prior knowledge of object appearance as a good starting point. While exploiting large-scale free web imagery, we introduce a sophisticated labour free method to construct a web dataset with good diversity in object appearance. After that, semantic relevance and distribution relevance are introduced and utilized in the proposed curriculum training scheme. Our end-to-end learning with the constructed web data achieves remarkable improvement across most object classes especially for the classes that are often considered hard in other works.
\end{abstract}

\section{Introduction}
\begin{figure}[t]
\includegraphics[width=\linewidth]{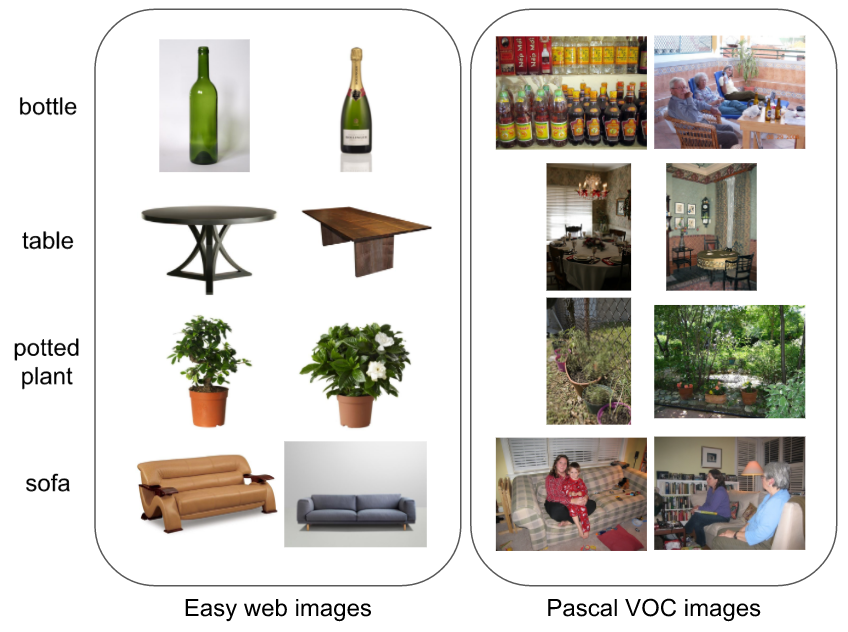}
\caption{Easy web images and VOC images. Web images have clean background while VOC images are more difficult with cluttered instances and complicated background. }
\label{fig:web_easy_voc_hard}
\end{figure}

With the rapid growth of computational power and dataset size and the development of deep learning algorithms, object detection, one of the core problems in computer vision, has achieved promising results \cite{ren2015faster,redmon2016you,liu2016ssd,li2016r}. However, state-of-the-art object detection methods still require bounding box annotations which cost extensive human labour. To alleviate this problem, weakly supervised object detection approaches \cite{wang2014weakly,bilen2014weakly,bilen2015weakly,bilen2016weakly,kantorov2016contextlocnet,cinbis2015weakly,tehattention} have attracted many attentions. These approaches aim at learning an effective detector with only image level labels, so that no labour-extensive bounding box annotations are needed. Nevertheless, as objects in common images can appear in different sizes and locations, only making use of image level labels are often not specific enough to learn good object detectors, and thus the performance of most weakly supervised methods are still subpar compared to their strongly supervised counterparts, especially for small objects with occlusions, such as ``bottle" or ``potted plant". As shown in Figure \ref{fig:web_easy_voc_hard}, images containing small objects or with very complicated contexts are hard to learn. In contrast, images containing a single object with very clean background provides very good appearance priors for learning object detectors. Particularly, for these easy images, the difficulty of localizing the objects is much lower than complicated images. With correct localization, the appearance model can be better learned. Therefore, easy images can provide useful information of the object appearance for learning the model for more complicated images. Unfortunately, such easy images are rarely available in object detection datasets, such as PASCAL VOC or MS COCO, as images in these multi-object datasets usually contain cluttered objects and very complicated background. On the other hand, there are a large number of easy web images available online and we can exploit these web images for the weakly supervised detection (WSD) task.

However, to construct a suitable auxiliary dataset and appropriately design an algorithm to utilize the knowledge from the dataset are non-trivial tasks. In this paper, we intend to provide a practical and effective solution to solve both problems.

Specifically, as various image search engines like Bing, Google, Flickr provide access to freely available web data of high quality images. Recent researches \cite{divvala2014learning,chen2015webly,xiao2015learning,xu2015augmenting,krause2016unreasonable} have already utilized these large-scale web data in various vision tasks. However, as object detection tasks impose specific requirements for auxiliary web data, we need to carefully design a labour-free way to obtain suitable images for the task.

First of all, when constructing the web dataset, we need to consider the relevance of web images in order to effectively transfer the knowledge of easy web images to the target detection dataset. In this paper, we break down this relevance into two parts, namely semantic relevance, which refers to the relevance between web images and the target labels, and the distribution relevance, which refers the relevance between web images and target images. As we will shown in later section, the semantic relevance focuses on a larger picture in the semantic space, while the distribution relevance measures more fine-grain differences in the feature distributions. To give an example, for category ``chair", the semantic relevance measures whether a certain web image is ``chair" or not, and the distribution relevance measures whether this web image lies on the manifold formed by the specific ``chairs" in the target dataset.  

Secondly, apart from the relevance problem, we also need to consider the diversity of the web images. As sub-categories, poses as well as backgrounds are crucial for the success of object detection, and thus our web images should not only be easy and related to the target dataset, but also contain a variety of different images even for the same category. With single text query, commonly used image search engines are not able to produce images with large intra-category diversity, especially in top ranked results. Therefore, inspired by \cite{divvala2014learning}, which uses ngram to retrieve the fine-grained dataset, we propose a multi-attribute web data generation scheme to enhance the diversity of web data. Specifically, we construct a general attribute table with common attributes that can easily be propagated to other target datasets as well. With the attribute table, we are able to build a hassle-free web dataset with proper category-wise diversity for the coarsely labeled dataset.

Once we have an appropriate web dataset, we need to consider how to transfer the knowledge from the easy web images to more complex multi-object target datasets. During the recent years, easy web images have been used in other weakly supervised tasks, such as weakly supervised segmentation~\cite{wei2016stc}. To the best of our knowledge, we are the first work bringing in web images for improving the weakly supervised object detection task.

Inspired by curriculum learning~\cite{bengio2009curriculum}, we propose a simple but effective hierarchical curriculum learning scheme. Specifically, with the hierarchical curriculum structure, all web images are considered easier than target images, which we refer as the first level of curriculum, followed by the second level of curriculum that includes all target images. 
Extensive experimental results show that our constructed web image dataset and the adopted curriculum learning can significantly improve the WSD performance.

\section{Related Work}
\label{related}
Our work is related to several areas in computer vision and machine learning.

\textbf{Weakly Supervised Object Detection (WSD)}: Traditional WSD methods like \cite{cinbis2015weakly} address this problem with multiple instance learning (MIL)~\cite{dietterich1997solving}, which treats each image as a bag and each proposal/window in the image as an instance in the bag. A positive image contains at least one positive instance whereas a negative image contains only negative instances. Since MIL approaches alternate the processes between selecting a region of objects and using the selected region to learn the object appearance model, they are often sensitive to initialization and often get stuck in local optima. \cite{bilen2016weakly} proposed a two-stream CNN structure named WSDDN to learn localization and recognition in dedicated streams respectively. These two streams share the common features from the earlier convolutional layers and one fully connected layer. It learns one detection stream to find the high responsive windows and one recognition stream to learn the appearance of the objects. In this way, the localization and recognition processes are decoupled. Similarly, \cite{kantorov2016contextlocnet} also uses the two stream structure and involves the context feature in the localization stream.

In this research, we use WSDDN \cite{bilen2016weakly} as an example to evaluate our learning method. Since WSDDN separates recognition and localization into two individual streams, it introduces additional degrees of freedom while optimizing the model, and hence it is hard to train at the early stage. It is also sensitive to initialization. Thus, in this work we propose to explicitly provide good initialization during the training process in an easy-to-hard manner. Note that although we utilize WSDDN as our baseline, our learning scheme is general and can be applied to other WSD methods as well.

\textbf{Curriculum Learning}: Our work is inspired by curriculum learning \cite{bengio2009curriculum} scheme. Curriculum learning was initially proposed to solve the shape recognition problem, where the recognition model is first trained to recognize the basic shapes and then trained on more complicated geoshapes. Recently, Tudor et al.~\cite{tudor2016hard} used this easy-to-hard learning scheme in MIL problem but mainly focused on learning a model to rank images with difficulty that matches the human perspective. In our work, we propose a hierarchical curriculum scheme that incorporates easy web images in early training stage to provide prior knowledge for the subsequent training on complicated images.

\textbf{Learning from Weak or Noisy Labels}: This paper is also related to those works on learning from weak or noisy labels \cite{divvala2014learning,reed2014training,chen2015webly,fu2015relaxing}. In \cite{divvala2014learning}, they proposed a  classifier-based cleaning process to deal with the noisy labels. They first train a classification model on images with higher confidence and then use this model to filter the outliers in the rest of images. Later, with incorporation of CNN, novel loss layer is introduced to the deep network in \cite{reed2014training}. In \cite{chen2015webly}, web images are separated into easy images (Google) and hard images (Flickr). They build a knowledge graph on easy web images and use the graph as a semantic constraint to deal with the possible label-flip noises during training of harder web images. Similarly, \cite{fu2015relaxing} learns the mutual relationship to suppress the feedback of noises during the back propagation. These works emphasize their methods to lessen the impact by outliers during the training process. In our work, apart from the outliers, we also consider distribution mismatch problem since we acquire web data that are from completely different information source with discrepant distribution compared to target dataset.


\section{Approach}
\label{approach}
In this part, we introduce the methodology on constructing the web dataset and the hierarchical curriculum learning to transfer the knowledge of web images to target dataset. We will use state-of-ther-art weakly supervised objection algorithm WSDDN~\cite{bilen2016weakly} as an example to show the effectiveness of our scheme. Note that our scheme is general and can be adapted to any other available algorithms if necessary.

\subsection {WSDDN}
 \label{sec:wsddn}
We first introduce weakly supervised deep detection network, or WSDDN~\cite{bilen2016weakly}, which is utilized as baseline for our experiments. WSDDN provides an end-to-end solution that breaks the cycle of training of classification and localization alternatively by decoupling them into two separate streams. 

\begin{figure}[t]
\includegraphics[width=\linewidth]{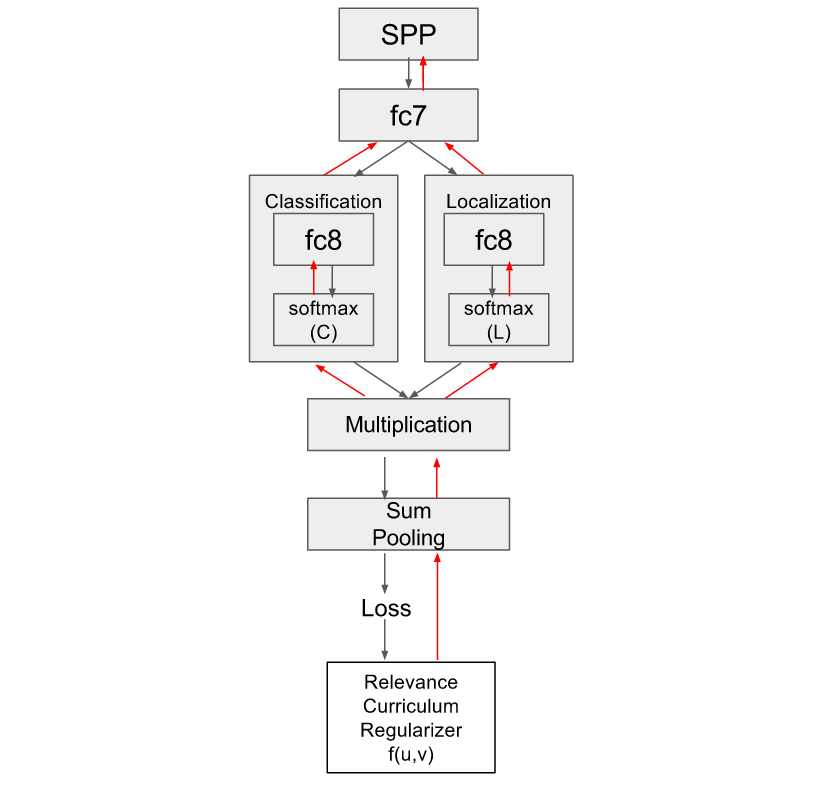}
\caption{WSDDN with relevance curriculum regularizer. The relevance curriculum regularizer suppresses backpropagation from samples which do not fit in the relevance region and curriculum region. }
\label{fig:wsddn_w_curr_rel}
\end{figure}

Particularly, WSDDN replaces the last pooling layer with spatial pyramid pooling layer \cite{lazebnik2006beyond} to obtain SPP feature of each region of interest (RoI). As shown in Figure \ref{fig:wsddn_w_curr_rel}, the SPP features are passed to a classification stream and a localization stream which individually learns the appearance and location of the objects. In the classification stream, the score for each RoI from \(fc8\) layer is normalized across classes to find the correct label of RoIs. In the localization stream, the scores of all RoIs are normalized category-wise to find most respondent RoIs for each category. Then the probability outputs from both softmax layers are multiplied as the final detection scores for each RoI. Finally, detection scores of all RoIs are summed up to one vector as the image level score to optimize the loss function (\ref{eq:wsddn_loss}).

\begin{equation}\label{eq:wsddn_loss}
L(y_{ci}, x_{i} | w) = -log(y_{ci}(\Phi_{c}(x_{i}|w)-\dfrac{1}{2})+\dfrac{1}{2})
\end{equation}

In the binary log loss function \(L(y_{ci}, x_{i} | w)\) , \(x_{i}\) is the input image i, and \(y\) is the binary image level label where \(y_ci=\{-1, 1\}\) for class $c$ in image $i$. Output from the last sum pooling layer is denoted as \(\Phi_{c}^y(x_{i}|w)\) which is a vector in range of $0$ to $1$ with the dimension equal to number of category. For each class $c$, if the label \(y_{ci}\) is $1$, \(L(y_{ci}, x_{i} | w) = -log(p(y_{ci}=1))\) and if \(y_{ci}\) is $-1$, \(L(y_{ci}, x_{i} | w) = -log(1-p(y_{ci}=1))\).

\subsection {Constructing Multi-Attribute Web Dataset}
 \label{sec:expand_condense}
In this section, we describe our method to construct a diversified and robust web dataset by introducing an expand-to-condense process. Specifically, we first introduce multiple attributes on top of the given target labels when crawling for web images to improve the generalization ability of the obtained dataset. Then we introduce both semantic relevance and distribution relevance to condense the dataset by filtering out irrelevant images.

\subsubsection{Expand to Diversify}
Free web images are abundantly available and accessible. Many image search engines can provide high quality images by searching the object names, such as Google, Flickr and Bing. In our preliminary study, we observe that images searched from Bing are generally easier than images from other search engines. Since easier images are intuitively better for learning object appearance, we choose Bing as the search engine to crawl web images. 

However, for most search engines, we observed that if we just use the given target labels as keywords, the resulting images are very similar in object appearances, poses or sub-categories. Moreover, the number of good quality images returned per query is very limited and lower ranked images are generally very noisy and unrelated to the queries.

\begin{table}
\small
\centering
\setlength\tabcolsep{4pt}
\caption{Attribute table.}\label{tb:attr_table}
\begin{tabular}{c|ccc}

\hline
Category&Viewpoint&Pose&Habitat\\
\hline
\hline
\makecell{aeroplane; bicycle;\\ boat; bus; car; \\motorbike; train; \\chair; diningtable; sofa } & \makecell{front view;\\side view }& -- & --\\
\hline
bird&\makecell{front view;\\side view} & -- &\makecell{water;\\ sky}\\
\hline
cat; dog &\makecell{front view;\\side view} & \makecell{sitting;\\walking;\\jumping} & --\\
\hline
cow; sheep& \makecell{front view;\\side view} &walking;& --\\
\hline
horse &\makecell{front view;\\side view} &\makecell{walking;\\jumping}& --\\
\hline
person &\makecell{front view;\\side view} &\makecell{sitting;\\standing;\\walking}& --\\
\hline
\end{tabular}
\end{table}

\begin{figure*}[t]
\includegraphics[width=\linewidth]{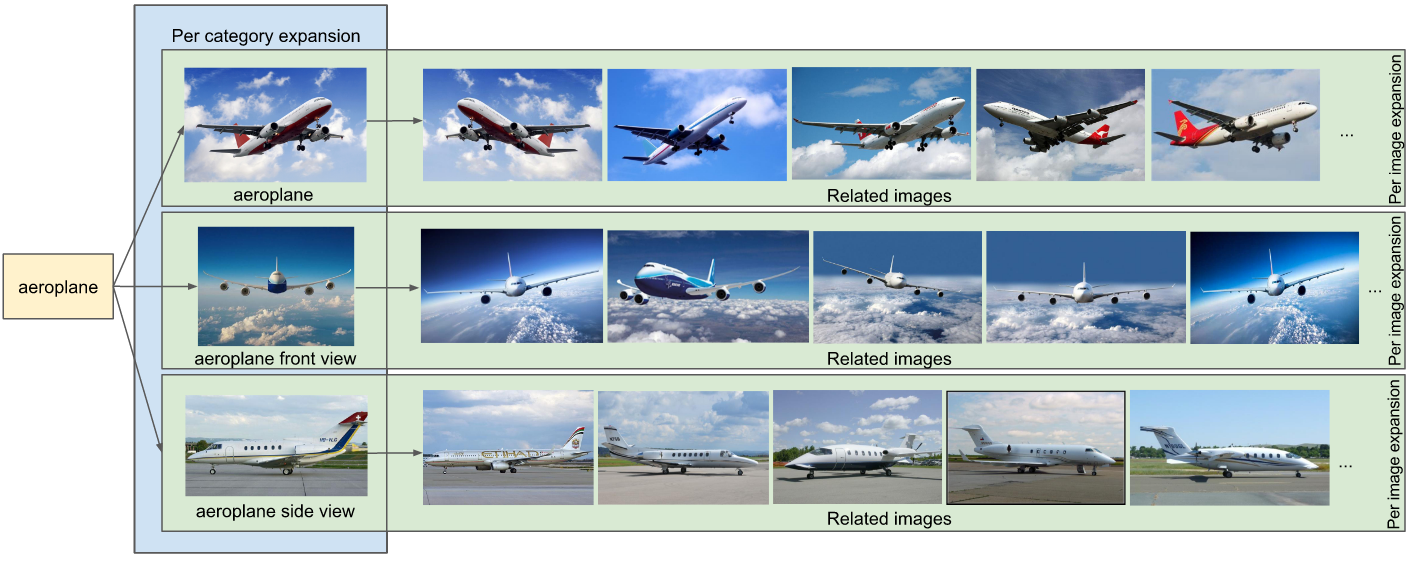}
\caption{Multi-attribute related dataset. Aeroplane category is expanded with multi-view attributes including front view and side view. Each multi-attribute web image is then expanded by the related images obtained from Bing image search engine.}
\label{fig:attr_rel}
\end{figure*}

To solve the problem of lacking diversity as well as limited number of high quality images, we introduce multiple attributes to each category. Based on the general knowledge of object detection, we define a set of attributes in three general aspects: namely viewpoints, poses or habitats of the objects. 

First of all, adding viewpoint attributes such as ``front view" and ``side view" not only provides extensive amount of high quality images for artificial objects like ``aeroplane", ``car" and ``bus", but also enhances the appearance knowledge of these objects, which will eventually make the detector more robust. Note that for categories without clear discrepancy between front view and side view such as ``bottle" and ``potted plant", as well as flat objects like ``tv monitor", we do not include these attributes. Secondly, for animals like ``cat" and ``dog", we add pose attributes. As their appearances vary significantly in different poses, adding such attributes will also be beneficial towards more robust detector. In particular, we add poses such as  ``sitting", ``jumping" and ``walking" to these animal categories. Last but not least, for category ``bird" which resides in different habitats, we add habitat attributes of ``sky" and ``water". The set of attributes is summarized in Table~\ref{tb:attr_table}. Note that following the same spirit, the table can be easily expanded to other categories.

Moreover, to overcome the limitation of limited available clean images in the top ranking, we also crawl related images. Related images are the images retrieved with similar visual appearance by using each of the previously retrieved top ranked images as query to the search engine. These related image can expand the size of the web dataset by more than 20 times and also introduce more variations to the dataset. Fig.~\ref{fig:attr_rel} illustrates the process of expanding the dataset by the multi-attribute per-category expansion and the per-image expansion.

\subsubsection{Condense to Transfer}
\label{sec:condense}
Once we obtain a large scale web image dataset, we are facing with the relevance problem. As free web data often contain many noisy images, to effectively make use of these web images, we need to analyse the image relevance to condense the noisy data. In this paper, we break down the image relevance to two parts: semantic relevance and distribution relevance. In detail, semantic relevance indicates whether a image contains the correct objects and distribution relevance measures how well a web image matches the the distribution of the target dataset.

Firstly, to measure the semantic relevance, we train a web-to-web outlier detector to find images with wrong labels in the web dataset. Specifically, we select top 80 images from queries of each target label and top 20 images from queries of each attribute + label combination. As we only use high ranked images as seed images, the ``cleanness" of the images can be guaranteed, and thus we are able to learn a more robust outlier detector. 

\begin{figure}
\includegraphics[width=\linewidth]{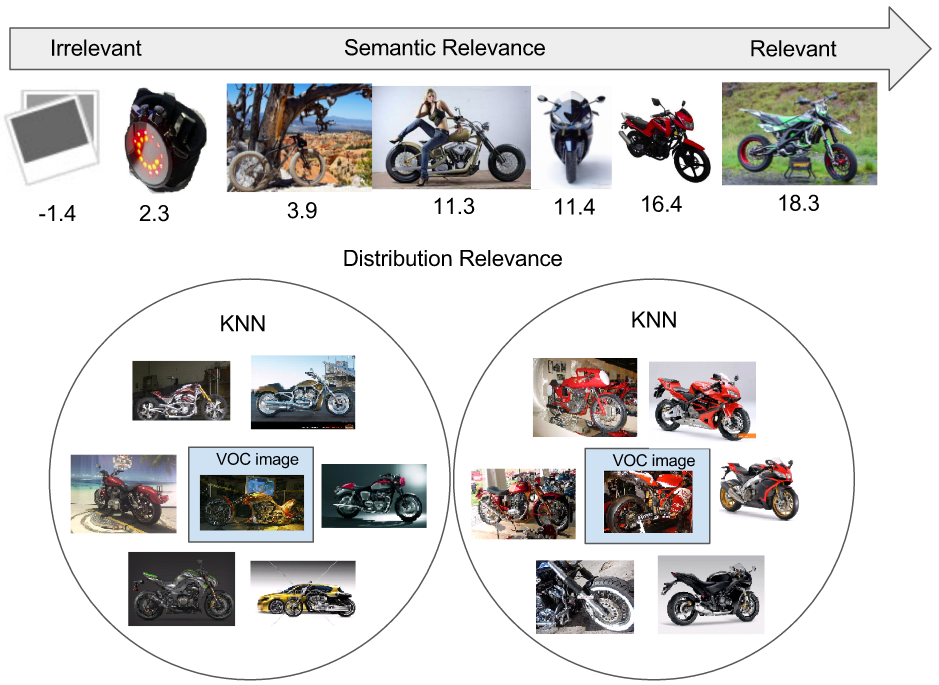}
\caption{Illustration of relevance metrics including semantic relevance and distribution relevance. Top: semantic relevance by the scores from web-to-web classifier for motorbike images, where non-meaningful images have negative scores, outliers with wrong objects have very low scores, and images containing correct objects have high scores.
Bottom: distribution relevance by $k$ nearest neighbors of each motorbike image in VOC dataset, where images in the neighborhood of VOC images with small feature distances are considered relevant to the target dataset.}
\label{fig:relevance}
\end{figure}

The outlier detector is trained iteratively with the expansion of the seed images. Similar to the idea of active learning, we train a CNN classifer with softmax loss with the seed images. Then it is applied to the whole set of the web images. The highly confident positive samples are then used as the second batch of training images for next iteration. After a few iterations, the classification scores from the final stabilized model are used to measure semantic relevance. As shown in Figure \ref{fig:relevance}, our model can provide very solid semantic relevance measurement. Most of the non-meaningful images have negative scores, outliers with wrong objects have very low scores and images with correct objects have high scores.

Secondly, since semantic relevance condenses images purely based on their semantic meaning regardless of the distribution matching with target dataset, we also consider the distribution relevance for more fine-grain measurements. To align the diversified web dataset into the distribution of target dataset, we search in the neighborhood of the target dataset to find similar web images. Particularly, for each single-label image in the target dataset, we select $k$ nearest web images in the feature space. The distance between images is defined as the Euclidean distance between their corresponding CNN features. Specifically, we use the L2 normalized \(fc7\) feature from a pretrained vgg-f model with PCA dimension reduction to represent each image. As shown in Figure \ref{fig:relevance}, our method is capable to extend the target dataset with web images having very similar object appearances and poses.  


We expect both relevance metrics to be effective for this task since it is intuitive to eliminate noises and unrelated data during the training. Nevertheless, our experiment result shows that matching the web data to target distribution is not as helpful as using a clean but diversified web dataset.

\subsection {Relevance Curriculum Regularizer}

Incorporating a good quality web dataset to the target dataset does not automatically guarantee better performance. Based on our experiments, we find out that simply appending these web images to target dataset is unhelpful or even harmful. These easy web images could lead to skew training models due to the distribution misalignment problem of the two datasets.

Therefore, instead of simply appending web data to target dataset, we propose a hierarchical curriculum structure. Specifically, we first consider a coarser curriculum with web images as easy and all target images as hard. If necessary, we could also add a fine curriculum to each dataset for full curriculum learning. Moreover, in addition to the normal curriculum or self-paced learning~\cite{jiang2015self}, we also consider adding an extra relevance term. As an analogy, we could consider web images as extracurricular activities. In order to help students with their learning, extracurricular activities need to be relevant to the course, in the same way that we should learn from easy images and relevant images.

In particular, to incorporate both curriculum and relevance constraints in training, we propose a relevance curriculum regularizer to the base detection structure:
\begin{equation}\label{eq:energy_curr_rel_term}
E(w) = \sum_{i=1}^{n}\sum_{c=1}^{C}L(y_{i} , x_{i} | w) \cdot f(u_{i},v_{i}),
\end{equation}
\begin{equation}\label{eq:curr_rel_term}
f(u_{i},v_{i}) = \sigma (u_{i}) \cdot \psi (v_{i}),
\end{equation}
where \(u_{i}\) is the relevance variable indicating whether the training sample is relevant as discussed in \ref{sec:expand_condense}. \(v_{i}\) is the curriculum regulation variable which indicates difficulty score of each image. \(\sigma\) is the relevance region function that only relevant samples can be learned every epoch. If a sample is in the relevance region, the value of \(\sigma(u)\) is $1$ and otherwise $0$. \(\psi\) is the curriculum region. It controls the pace of learning that allows only easy samples to be learned at early stage and gradually adding harder samples along the training process. If the difficulty score of sample image is within the curriculum region, \(\psi(v)\) is $1$ and otherwise, \(\psi(v)\) is $0$. As described previously, we implemented a hierarchical curriculum, where \(\psi(v)\) for all web images are consider as $1$ first, then we gradually expand it to include target images.

\section {Experiments}
\label{exp}
In this section, we evaluate the effectiveness of our proposed weakly supervised object detection.

\subsection {Baseline Model \& Setting}


For experimental setting, similar to the original WSDDN work, we use Edgebox~\cite{zitnick2014edge} as the proposal method to generate around 2000 bounding boxes. To train the network, we use the vgg-f model pretrained on ImageNet as the initial model. For fairness, our results are compared with the baseline method trained on vgg-f as well. 
We evaluate our method on PASCAL VOC2007 and VOC2012 datasets with 20 object categories. During the training, we use only image-level labels of the training images. The evaluation metric is the commonly used detection mAP with IoU threshold of 0.5.

\subsection {Results Regarding Curriculum Learning}
\begin{figure}
\includegraphics[width=\linewidth]{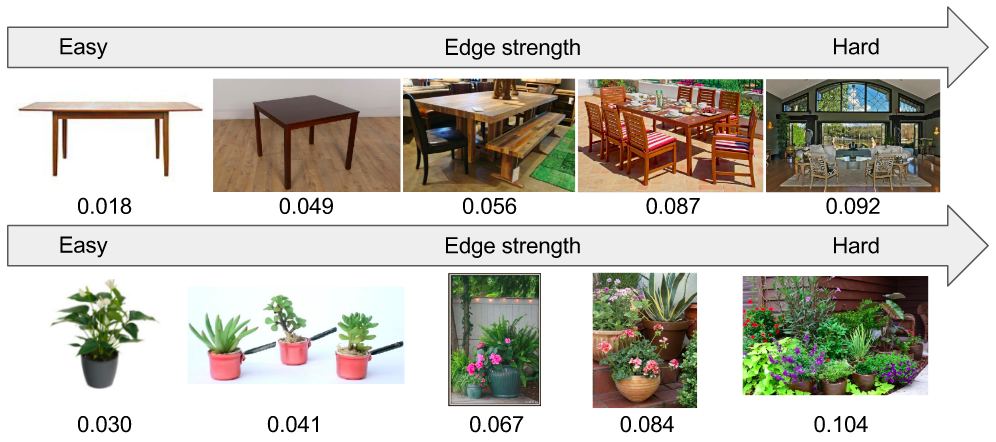}
\caption{Curriculum metric by the mean edge strength for web images (Top: table / Bottom: potted plant). The mean edge strength can reasonably represent the difficulty of images. Images with clean background and single object usually have small mean edge strength and images with complicated background and cluttered objects usually have large edge strength.}
\label{fig:edge_strength}
\end{figure}

We first evaluate the effectiveness of applying the curriculum learning method on PASCAL VOC 2007 trainval set itself, without using our web data. The curriculum is designed by the ranking of the mean edge strength of each image. The mean edge strength of an image is defined as the number of edge pixels over the total number of pixels. This is a simple yet intuitive method because images with more edges tend to have more complicated background or contain more cluttered objects, and thus it is reasonable to consider them as hard samples. Fig.~\ref{fig:edge_strength} gives some examples, which show that the mean edge length represents the relative difficulty of the images well.

\begin{table}
\centering
\caption{Results of the detection mAP on VOC2007 test set by using the curriculum regularization term to train VOC 2007 trainval set. } \label{tb:CurrWSDDN}
\begin{tabular}{| l | r | }
\hline
Methods & mAP\\ \hline \hline
WSDDN (baseline)    & 33.9 \\ \hline
CurrWSDDN   & \textbf{35.5} \\ \hline
\end{tabular}
\end{table}

Specifically, we use the classical LoG edge detector to detect edges. For each curriculum region, we add $\frac{1}{5}$ of more difficult images from each category. This is to balance the number of positive samples from each category in every iteration. In this way, the curriculum consists of five overlapped regions with gradually increased image complexity. Table~\ref{tb:CurrWSDDN} shows the detection result (`CurrWSDDN') of applying the curriculum regularization term to train VOC 2007 trainval set only, compared with the result of the baseline (`WSDDN'). We can see that using curriculum learning on VOC 2007 training images alone already improves the performance. This suggests that for training weakly supervised object detector, it is beneficial to train the network in an easy-to-hard manner. Note that the baseline WSDDN result is obtained by running the original WSDDN codes released in Github with the same setting\footnote{https://github.com/hbilen/WSDDN}, which is slightly different from the result of 34.5 reported in~\cite{bilen2016weakly}.

\begin{table}
\centering
\caption{Results of the detection mAP on VOC2007 test set by using our constructed Bing dataset or the `Flickr clean' dataset as easy images and VOC images as hard images for easy-to-hard training. } \label{tb:WebETH}
\begin{tabular}{| l | r | }
\hline
Web dataset & mAP\\ \hline
\hline
WebETH(Flickr clean)& 35.5 \\ \hline
WebETH(Bing)& \textbf{36.0} \\ \hline
\end{tabular}
\end{table}

\begin{table}
\centering
\caption{Results of the detection mAP on VOC2007 test set with different relevance metrics and using our constructed Bing dataset with the easy-to-hard training. }\label{tb:WebRelETH}
\begin{tabular}{| l | r | }
\hline
Transfer metrics & mAP\\ \hline
\hline
WebRelETH(Dist-Rel)& {35.9} \\ \hline
WebRelETH(Semantic-Rel)& \textbf{36.8} \\ \hline
\end{tabular}
\end{table}

\subsection{Results Regarding Constructed Web Dataset}
We now evaluate the usefulness of our constructed web image dataset for WSD. 
As mentioned in Section \ref{sec:expand_condense}, we construct a web image dataset of 34k images using Bing image search engine with attributes and related images. Considering that many selected web images are of high resolution, which causes huge complexity in the proposal generation process, we resize the longer side of all images to 600 pixels and keep the aspect ratios. 
We treat all web images as easy images and all VOC images as hard images. Simple web images are trained first followed by more complicated VOC images. Table~\ref{tb:CurrWSDDN} shows the detection result of our method `WebETH(Bing)' that exploits our constructed Bing dataset and trains the network in an easy-to-hard manner. Comparing Tables~\ref{tb:CurrWSDDN} and~\ref{tb:WebETH}, we can see that our method `WebETH(Bing)' significantly improves the baseline `WSDDN', increasing mAP from 33.9\% mAP to 36\%, and also outperforms the VOC curriculum method `CurrWSDDN'.

We also conduct experiments on another publicly available web dataset, STC Flickr clean dataset~\cite{wei2016stc}, which contains 
more than 40k super clean images and has been proven to have good performance in generating good saliency maps to train weakly supervised segmentation networks. Surprisingly, by involving STC Flickr clean, although its result (see Table~\ref{tb:CurrWSDDN}) is much better than the baseline using only VOC images, it has no improvement over the VOC curriculum method `CurrWSDDN'. In contrast, using our noisy Bing dataset `WebETH(Bing)' beats both the VOC curriculum method `CurrWSDDN' and the Flickr clean dataset `WebETH(Flickr clean)'. This suggests that our approach of constructing a multi-attribute web dataset with large diversity is practically useful in this context. 
\begin{table*}
\setlength\tabcolsep{4pt}
\centering
\caption{Comparisons of the detection average precision results (\%) on VOC2007 test set with training on VOC2007 trainval set. We use vgg-f model pretrained on ImageNet. WSDDN results are obtained using the published code on Github with the same setting stated in~\cite{bilen2016weakly}. Results of other methods are from their papers.}\label{tb:results_2007}
\resizebox{\textwidth}{!}{%
\begin{tabular}{|c|cccccccccccccccccccc||c| }
\hline
&aero&bike&bird&boat&bottle&bus&car&cat&chair&cow&table&dog&horse&mbike&person&plant&sheep&sofa&train&tv&mean\\ \hline
\hline
Bilen et al. \cite{bilen2014weakly} &42.2&43.9&23.1&9.2&12.5&44.9&45.1&24.9&8.3&24.0&13.9&18.6&31.6&43.6&7.6&\textbf{20.9}&26.6&20.6&35.9&29.6&26.4\\
\hline
Bilen et al. \cite{bilen2015weakly}&46.2&46.9&24.1&16.4&12.2&42.2&47.1&35.2&7.8&28.3&12.7&21.5&30.1&42.4&7.8&20.0&26.8&20.8&35.8&29.6&27.7\\
\hline
Cinbis et al. \cite{cinbis2015weakly} &39.3&43.0&28.8&20.4&8.0&45.5&47.9&22.1&8.4&33.5&23.6&29.2&38.5&47.9&20.3&20.0&35.8&30.8&41.0&20.1&30.2\\
\hline
Wang et al. \cite{wang2014weakly}&48.8&41.0&23.6&12.1&11.1&42.7&40.9&35.5&11.1&36.6&18.4&35.3&34.8&51.3&17.2&17.4&26.8&32.8&35.1&45.6&30.9\\
\hline
Teh et al. \cite{tehattention}&48.8&45.9&37.4&\textbf{26.9}&9.2&50.7&43.4&43.6&10.6&35.9&27.0&38.6&48.5&43.8&\textbf{24.7}&12.1&29.0&23.2&48.8&41.9&34.5\\
\hline
ContextLocNet(contrastive S) \cite{kantorov2016contextlocnet}&\textbf{57.1}&52&31.5&7.6&11.5&55&53.1&34.1&1.7&33.1&\textbf{49.2}&42&47.3&56.6&15.3&12.8&24.8&\textbf{48.9}&44.4&47.8&36.3\\
\hline
\hline
WSDDN \cite{bilen2016weakly} &41.8&\textbf{57.7}&31.8&16.2&9.2&59.2&53.0&39.1&3.6&34.6&14.2&33.5&50.2&53.5&9.8&15.6&\textbf{37.3}&21.0&\textbf{53.1}&43.3&33.9\\
\hline
\hline
WSDDN(Flickr clean only)&31.4&26.6&22.0&10.0&1.5&43.0&38.1&36.6&1.7&12.3&19.7&32.8&34.1&38.6&8.4&5.7&17.6&29.5&32.0&18.2&23.0\\
\hline
WSDDN(Bing rel only)&37.4&22.6&18.5&6.9&1.7&42.2&38.0&29.9&1.0&14.9&1.7&37.1&34.2&33.9&11.7&4.4&17.0&16.3&27.7&12.5&20.5\\
\hline
CurrWSDDN &40.4&54.6&28.2&15.4&10.4&57.4&53.0&\textbf{44.5}&1.2&35.3&30.9&41.5&51.3&53.0&11.6&16.3&34.5&39.0&46.0&45.0&35.5\\
\hline
WebRel(ours) &40.7&51.5&31.0&10.7&10.0&61.0&43.2&39.4&1.8&30.1&35.5&\textbf{46.4}&52.3&50.6&9.0&13.4&30.4&31.8&41.2&42.3&33.6\\
\hline
WebETH(ours) &40.2&51.6&33.3&13.5&\textbf{13.0}&\textbf{62.8}&\textbf{54.5}&38.7&\textbf{11.8}&34.8&25.1&42.2&50.5&55.3&13.1&19.0&31.4&34.6&49.3&44.6&36.0\\
\hline
WebRelETH(ours) &44.4&52.1&\textbf{38.1}&10.2&12.3&61.5&54.4&33.5&7.6&37.2&30.2&37.6&\textbf{55.4}&\textbf{57.3}&9.1&18.3&35.9&43.0&47.6&\textbf{50.0}&\textbf{36.8}\\
\hline
WebRelETC(ours) &38.9&52.4&33.4&11.2&10.5&59.9&53.8&36.4&3.0&\textbf{38.5}&41.8&38.8&53.9&56.0&11.9&18.9&35.1&43.2&46.2&47.2&36.6\\
\hline
\end{tabular}
}
\end{table*}

\begin{table*}
\setlength\tabcolsep{4pt}
\centering
\caption{Results of the detection average precision (\%) on VOC2012 test set with training on VOC2012 training set. We use vgg-f model pretrained on ImageNet. WSDDN results are obtained using the published code on Github with the same setting stated in~\cite{bilen2016weakly}.}\label{tb:results_2012}
\resizebox{\textwidth}{!}{%
\begin{tabular}{|c|cccccccccccccccccccc||c| }
\hline
&aero&bike&bird&boat&bottle&bus&car&cat&chair&cow&table&dog&horse&mbike&person&plant&sheep&sofa&train&tv&mean\\ \hline
\hline
WSDDN \cite{bilen2016weakly} &53.4&53.2&36.2&7.9&16.4&57.2&35.3&24.8&6.5&29.0&13.7&31.1&47.1&57.2&11.0&18.9&28.6&19.4&\textbf{42.3}&39.6&31.4\\
\hline
\hline
WebRelETH (ours)&\textbf{57.6}&55.1&\textbf{38.5}&\textbf{8.6}&20.4&\textbf{59.4}&\textbf{36.4}&33.6&\textbf{14.0}&\textbf{34.8}&21.7&\textbf{39.4}&\textbf{51.3}&\textbf{62.8}&11.5&\textbf{19.2}&\textbf{30.2}&\textbf{23.9}&41.2&\textbf{44.5}&\textbf{35.2}\\
\hline
WebRelETC (ours)&55.6&\textbf{56.2}&35.3&7.4&\textbf{20.5}&55.6&32.6&\textbf{34.8}&9.7&32.9&\textbf{32.1}&34.6&48.4&61.6&\textbf{15.5}&18.9&27.3&15.7&41.2&43.5&34.0\\
\hline
\end{tabular}
}
\end{table*}

\begin{figure*}[h!]
\centering
\includegraphics[width=0.8\linewidth]{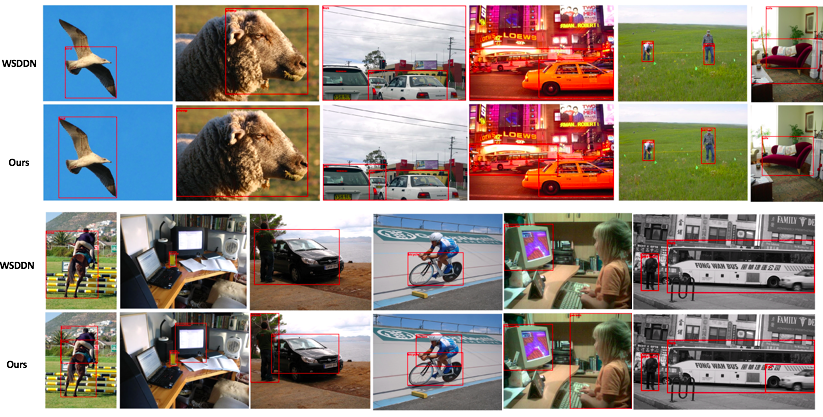}
\caption{Visual results of WSDDN and our best model (WebRelETH). Our model can refine the bounding boxes as shown in the top two rows. Missing objects in the original model can also be detected in some test images as shown in the bottom rows. }
\label{fig:results}
\end{figure*}

\subsection {Results Regarding Relevance Metrics}
Here we conduct experiments to study the effectiveness of using semantic relevance and distribution relevance. Fig.~\ref{fig:relevance} gives some examples of the two relevance metrics. For the semantic relevance, we use the classification scores by the outlier detector described in Section~\ref{sec:condense}, whose values vary from negative to more than 20. We set a semantic relevance threshold of 8 so that web images with scores lower than 8 are excluded. This prevents from mixing in noisy images without target objects into the early stage of training. For the distribution relevance, its relevance region includes web images which are members of top $k$-th nearest neighbors of one of VOC images, as illustrated in Fig.~\ref{fig:relevance}.

Table~\ref{tb:WebRelETH} shows the results using the two relevance metrics. We can see that with the semantic relevance, the detection result increases from 36.0\% to 36.8\%, whereas the kNN based distribution relevance gives a slightly lower result, which suggests that similar images might not be always preferred. As a non-convex optimization problem, the training of WSD tends to drift to optimize small clusters of training samples. Although additional training instances with a similar distribution can help achieve lower training loss, it is not as helpful as involving new training samples with larger diversity, which leads to better generalization ability. This may also explain why STC Flickr clean dataset is not so helpful since the images in the Flickr clean dataset also have a similar distribution as VOC dataset.

\subsection {More Comparison Results}

Table~\ref{tb:results_2007} lists out the per-category average precision results of different WSD methods on VOC2007 test set with training on VOC2007 trainval set. It can be seen that compared with other existing WSD methods, the baseline method WSDDN achieves reasonably good performance. We would like to point out that our list in Table~\ref{tb:results_2007} might not be exhaustive since there might be some very recent WSD methods that report better performance. Since our solution is general, which can be added on top of any WSD baseline, it is more meaningful to evaluate our methods w.r.t the baseline.  

Based on WSDDN, we consider five variants: using only VOC images with the curriculum regularizer (CurrWSDDN), simply combining our web images with VOC images with the semantic relevance for training (WebRel), combining our web images with VOC images for easy-to-hard training (WebETH), combining our web images with VOC images with the semantic relevance for easy-to-hard training (WebRelETH), and combining our web images with VOC images with the semantic relevance for easy-to-curriculum training (WebRelETC), where we train easy web images first and then train VOC images in a more detailed curriculum.

The results of CurrWSDDN, WebETH and WebRelETH have been discussed previously w.r.t. Tables~\ref{tb:CurrWSDDN}, \ref{tb:WebETH} and \ref{tb:WebRelETH}, which demonstrate the effectiveness of the curriculum regularizer, the constructed web dataset, the proposed relevance metrics, respectively. For WebRel, its result is even worse than the baseline WSDDN, which suggests that it is not an effective way to simply combine data from two sources. In our case, a large number of easy images dominate the training so that the model cannot be well trained for hard samples. For WebRelETC, we expect that web images to have similar difficulty level but VOC images need to be partitioned in more levels of difficulty. We first train on easy web images and adopt five-level curriculum regions for VOC images. It is found that its average precision performance is slightly worse than WebRelETH. This suggest that it is not always good to further break down the higher level curriculum for every class if the lower-level curriculum of simple web images have been used. Overall, our WebRelETH achieves the best mAP of 36.8\%, outperforming the baseline by 2.9\%.

Table \ref{tb:results_2012} shows the experiment results for VOC2012. Our method also achieved up to 3.8\% improvement in this dataset. Similar to VOC2007, WebRelETH outperforms WebRelETC, although WebRelETC excels largely in ``dining table" by more than 10\%. Fig.~\ref{fig:results} gives some visual comparisons of the detection results using WSDDN and our best model (WebRelETH). It can be seen that our model can refine the bounding boxes (see the top two rows of Fig.~\ref{fig:results}), and missing objects in WSDDN can also be detected by our model in some test images (see the bottom rows of Fig.~\ref{fig:results}).

\section {Conclusion}
\label{conclusion}
This paper have addressed the two questions: how to construct a large, diverse and relevant web image dataset and how to use it to help weakly supervised object detection. Particularly, for constructing the web dataset, we introduced a sophisticated expand-to-condense process to first expand web data with attributes and related images and then condense the dataset with semantic relevance or distribution relevance. For helping the target dataset, we applied an easy-to-hard learning scheme. Extensive results have validated that our easy-to-hard learning with web data is effective and the multi-attribute web data do help in training a weakly supervised detector.


\clearpage
{\small
\bibliographystyle{ieee}
\bibliography{tqy}
}

\end{document}